# AN ANALYSIS OF THE METHODS EMPLOYED FOR BREAST CANCER DIAGNOSIS

Mahjabeen Mirza Beg[1], Monika Jain[2]

[1] B.Tech (4th year,) Electronics and Instrumentation Engineering

Galgotias College of Engineering & Technology, Gr. Noida.

Email: mirza.mahjabeen@yahoo.in

[2] Head of Department Electronics & Instrumentation Engineering

Galgotias College of Engineering & Technology, Gr. Noida.

Email: monikajain.bits@gmail.com

***Abstract:*** *Breast cancer research over the last decade has been tremendous. The ground breaking innovations and novel methods help in the early detection, in setting the stages of the therapy and in assessing the response of the patient to the treatment. The prediction of the recurrent cancer is also crucial for the survival of the patient. This paper studies various techniques used for the diagnosis of breast cancer. Different methods are explored for their merits and de-merits for the diagnosis of breast lesion. Some of the methods are yet unproven but the studies look very encouraging. It was found that the recent use of the combination of Artificial Neural Networks in most of the instances gives accurate results for the diagnosis of breast cancer and their use can also be extended to other diseases.*

## I. Introduction

Breast cancer is the second most fatal disease in women worldwide [1-4], the risk increasing with the age. Breast cancer affects not only women but also men and animals. Only 1% of all the cases are found in men. There are two types of breast lesions- malignant and benign. The Radiologists study various features to distinguish between the malignant tumor and benign tumor. 10%-30% of the breast cancer lesions are missed because of the limitations of the human observers [5, 6]. The malignant tumor is in many cases misdiagnosed and its late diagnosis reduces the chances of survival of the patient. Early and accurate diagnosis is essential for patient's timely recovery. Identifying the women at risk is an important strategy in reducing the number of women suffering from breast cancer.

Detecting the probability of recurrence of the cancer can save a patient's life. Conventionally, biopsy was used for the diagnosis, nowadays mammography, breast MRI, ultrasonography, BRCA testing etc. are done. When a number of tests are performed on a patient it becomes difficult for the medical experts to come to a correct conclusion and the screening methods produce false positive results. Thus smarter systems are required to decrease the instances of false positives and false negatives.

This paper reviews the existing/popular methods which employ the soft computing techniques to the diagnosis of breast cancer.

## II. Literature Survey

The Computer-Aided-Diagnosis has been proposed for the medical prognosis [7-9]. The fuzzy logic and Artificial Neural Network form the basis of the intelligent systems. There are several instances where the artificial intelligence is used for the diagnosis of the breast cancer. The methods have included many Artificial Neural Networks architectures such as Convolution Neural Network [10], Radial Basis Network [11], General Regression Neural Network [11], Probabilistic Neural Network [11], Resilient Back propagation Neural Network [12], and hybrid with Fuzzy Logic [13]. Most of the papers used MATLAB, a high performance and easy to use environment; for the diagnosis and classification of the breast cancer. In this paper [7] a supervised artificial neural network [14-16] was used to help classify the breast lesions into malignant and benign classes by processing the computer cytology images. Accuracy of trained neural network was found to be 82.21%. The ANN has been established as a robust system for the diagnosis of breast cancer [17].There is a complex relationship between different biomarkers which were identified for the diagnosis of this cancer [18], the MLP neural network was simulated for the diagnosis using four biomarkers (DNA ploidy, phase fraction (SPF),cell cycle distribution and the state of steroid receptors) and it was found that this method is better than previously used techniques like logistic regression[19].Different combinations of the biomarkers were applied to the MLP and it was concluded that DNA had no effect on the outcome thus it can be excluded from the prognosis. In this paper [20] the values of the features like clump thickness, uniformity of cell size, uniformity of cell shape, etc. are first normalized. The lower ranked features were removed using the information gain method and the higher ranked attributes were fed to the ANFIS (as shown in figure 1), which were processed and the accuracy of this method when applied to the Wisconsin Breast Cancer Diagnosis (WBCD) dataset was found to be 98.24% but no heed was paid to the computational time.

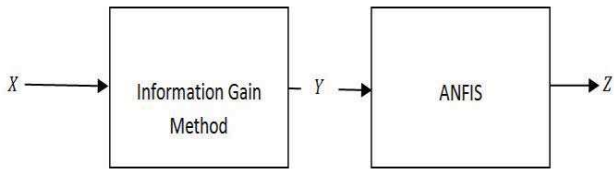

*Figure 1: General Structure of the Proposed Method*

The quality of the attributes in the information gain method was estimated by calculating the difference between the post probability and prior probability thereby reducing the number of features from nine to four. The figure 2 shows the ranking of the attributes using the *InfoGainAttributeVal* and the searching method *Ranker-T-1* using WEKA on WBCD dataset where WEKA is JAVA language machine learning software.

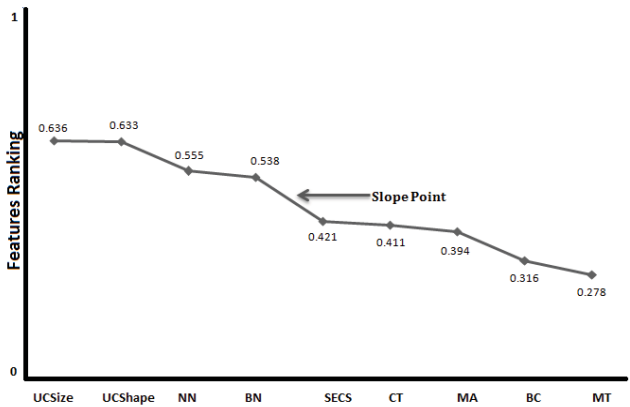

*Figure 2: Information Gain Ranking*

In the next stage a Sugeno Fuzzy Inference system (FIS) was built using the MATLAB FIS toolbox. The inputs were the four attributes with high ranks and the output were the two classes of tumor. The FIS contained 81 rules and it was loaded to the ANFIS for training and testing of the method. The structure of the ANFIS is shown in figure 3. Thus this method reduced the complexity of the problem.

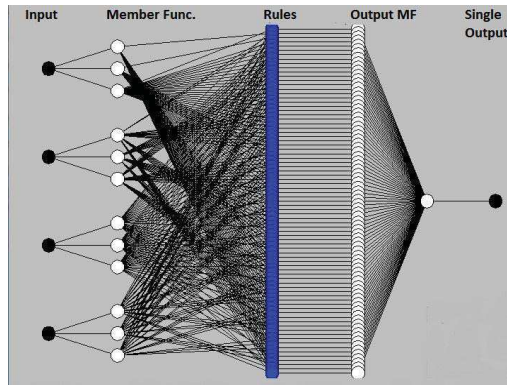

*Figure 3: AFIS Structure on MATLAB*

Modular Neural Networks were built by brute force ray tracing algorithm into small modules [21]. MNNs give better performance than the monolithic NNs, such as increased reliability, better generalization ability and faster performance. The application of ANN to the diagnosis can be divided into two parts- training and testing. To solve the problem of large dimensionality, all the attributes were divided into two parts, each part contained half the number of attributes, thus inserting modularity at attribute level and reducing the complexity of the problem. The limitations of the single neural networks were removed by using multiple neural networks. Back propagation neural network (BPNN) and radial basis function network (RBFN) were used for the training and testing of data; resulting into four modules. The modules gave the probability of occurrence of disease in the form of probability vector which had values between 0 and 1, where 0 denoted the absence of disease and 1 denoted the presence of disease. The weights associated with each module were real numbers set by the designer so as to maximise the network performance. The outputs of the modules were fed to the integrator which made the final diagnostic decision given by:

$O = o_1w_1 + o_2w_2 + o_3w_3 + o_4w_4$

Where $w_1+w_2+w_3+w_4=1$.

If the value of O was greater than 0.5 then it was classified as benign and if it was greater than 0.5 then malign. The experimental results were as shown in table 1.

*Table 1: Experimental Results*

| Module # | Methods | Attributes | Training accuracy | Testing accuracy | Time (sec) |
|---|---|---|---|---|---|
| 1 | BPA | 1-15 | 89.50% | 96.4% | 3.88 |
| 2 | RBFN | 1-15 | 94.75% | 96.44% | 0.25 |
| 3 | BPA | 16-30 | 91.50% | 94.67% | 3.82 |
| 4 | RBFN | 16-10 | 97.50% | 97.63% | .29 |
| - | MNN | 1-30 | 95.75% | 98.22% | 8.24 |
| - | BPNN | 1-30 | 91% | 96.44% | 5.58 |
| - | RBFN | 1-30 | 97.25% | 97.63% | .25 |

The paper demonstrated the better performance of the multiple neural networks over the monolithic neural networks. The approach can be extended to other large data sets.

A novel application specific instrumentation technique was designed by Mishra and Sardar [22] and it was used for the simulation of breast cancer diagnosis system using the ultra-wideband sensors. The problems with generic instrumentation systems are that the human interpreter is inevitable and is very costly; the ASIN removed both these problems. The UWB sensors used, remove the need for image reconstruction. The RBF based ANN was used to detect the presence of the tumor and the Finite difference time domain method was used for the simulation. The large differences between the tumor and other breast organisms help in its easy detection. The method though tested only on simulated dataset looks very promising as the correct detection rate was found to be very high, the cost of the system was reduced by many folds and the need for human expert was also removed. Jamarani, *et.al* developed and constructed a method which used the Wavelet Packet based neural network [23]. The micro calcifications correspond to high frequency thus the lower frequency bands were suppressed, the mammogram was divided into sub frequency bands and reconstructed using only the sub bands of high frequencies. The results from wavelets were fed to the ANN. The method was found to be 96%-97% accurate and the system successfully combined the intelligent techniques with the image processing thereby increasing the sensitivity of the diagnosis.

Sometimes, even after the primary treatment breast cancer can return. The prediction of the recurrent cancer is a very challenging task; reference [24] developed a method for the aforesaid. The conventional imaging (CI) with an accuracy of up to 20% or the complex and expensive methods like Magnetic Resonance Imaging (MRI) or Positron emission Tomography (PET) with an accuracy of 80% are used for such diagnosis thus this paper used the RBF, MLP and PNN for the same. The NN algorithm designed was found to be accurate but the PNN performed poorly. The MLP and RBF gave good performance but the performance of MRI and PET is very high. Renjie Liao; Tao Wan and Zengchang Qin [25] developed a CAD system for differentiating the benign breast nodules from the malignant nodules. The discrimination capability of the features extracted from the sonograms was tested by using the SVM (support vector machine), ANN and KNN (K-nearest neighbor) classifier. It was found that the SVM gave the greatest accuracy while the ANN had the highest sensitivity. The features extracted from the images were fed to the neural network [26]. The fuzzy co-occurrence matrix and fuzzy entropy method were used for features' extraction and the data was fed to feed-forward multilayer neural network to classify the biopsy images into three classes. The FCM though has small dimensions yet is more accurate than the ordinary co-occurrence matrix. The performance of the method was found to be better than the other conventional methods as the fuzziness of the data was also considered. The method gave 100% classification result but the typical co-occurrence matrix cannot attain accurate diagnosis. This paper [27] uses the Jordan Elman neural network approach on three different data sets. The Jordan-Elman NN differs from NN such that the feedback is from output layer to the input layer instead of the hidden layer as shown in figure 4. It was found that the approach can aid the medical experts in diagnosis to prevent biopsy.

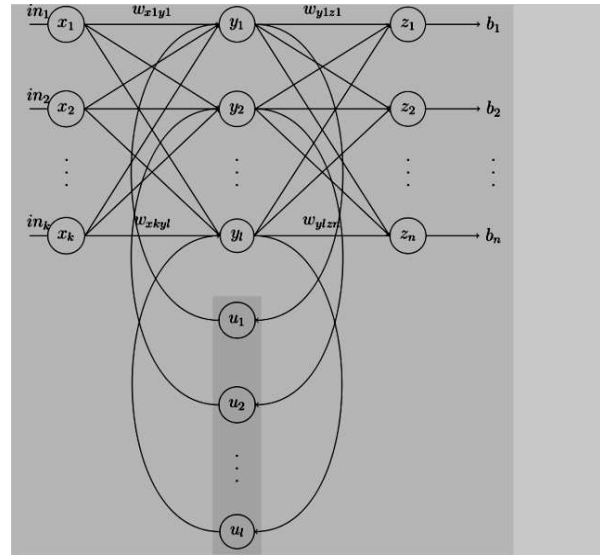

*Figure 4: Jordan-Elman Neural Network Structure*

The malignant cancer cell can be effectively diagnosed. The performance of the unsupervised and supervised neural network for the detection of breast cancer has been presented by Belciug *et.al* [28]. Only an unsupervised NN will help in assessing the medical expert in case of a patient with no previous diagnosis. The comparison of the diagnosis ability of the four types of NN models (MLP, RBF, PNN, and SOFM) was done. The SOFM is easy and it exploits its self-organizing feature, these are its advantages over the standard NNs. However there is scope of future work to assess this hypothesis. In [29] the back propagation algorithm is compared with the Genetic algorithm for the CAD diagnosis of breast cancer using the receiver-operating characteristics (ROC). The GA slightly outperformed the BP for training of the CAD schemes but not significantly. The GA is better used for the feature selection.

Most of the methods designed/used/tested in various papers use soft computing to identify, classify, detect, or distinguish benign and malignant tumors. Majorly all the methods used ANNs at some stage of the process or the other and different combinations of NNs were shown to give better results than the use of a single type of NN.

## III. Conclusions

The last decade has witnessed major advancements in the methods of the diagnosis of breast cancer. Only recently the soft computing techniques are being used, hence the body of study in this area is very less. The CAD systems reduce the false alarms. It was found that the use of ANN increases the accuracy of most of the methods and reduces the need of the human expert. The neural networks based clinical support systems provide the medical experts with a second opinion thus removing the need for biopsy, excision and reduce the unnecessary expenditure. The design of ANNs must be optimized according to a specific problem; simply using a generic ANN may reduce efficiency and lead to slow learning. The ANN, SVM, GA, and KNN may be used for the classification problems. Almost all intelligent computational learning algorithms use supervised learning. Supervised ANN outperforms the unsupervised network but in the case of a patient with no previous medical records the unsupervised ANN is the only solution. The RBFN due to their highly localized nature perform poorly when used for the classification problems. The accuracy of different architectures are in the order LVQ followed by CL, MLP and RBFN. Some of the methods can also be extended to other diseases. The ANN predominates but it is evident that other machine learning algorithms are also being developed. The accuracy of different methods on different dataset is compared in table 2.

*Table 2: Comparison of Accuracy of Different Methods*

| The approach | Dataset | Accuracy | Reference |
|---|---|---|---|
| SANE | WBCD | 98.7% | [2] |
| IGANFIS | WBCD | 98.24% | [20] |
| ASIN on observation from UWB sensors | Simulated data | 98% | [22] |
| SVM | Harbin Institute of Technology and the Second Affiliated Hospital of Harbin Medical University. | 86.92% | [25] |
| ANN | | 86.60% | |
| KNN | | 83.8% | |
| fuzzy co-occurrence matrix concept | diagnosed breast-tissue sample images | 100% | [26] |
| Xyct system using Leave One Out method | WBCD Visually extracted | 91% | [30] |
| | WBCD Digitally extracted | 90% | |
| ANFIS | WBCD | 59.90% | [31] |
| FUZZY | WBCD | 96.71% | [32] |
| SIANN | WBCD | 100% | [33] |
| JENN | WBCD | 98.75% | [27] |
| | WDBC | 98.25% | |
| | WDPC | 70.725% | |

The test accuracies of some of the popular and efficient methods are compiled in table 2. The analysis showed that the diagnosis when used fuzzy co-occurrence matrix for features' extraction gave 100% accuracy and the SIANN method also gave 100% accuracy.